%% file: main.tex
\pdfoutput=1

\documentclass[11pt]{article}

\usepackage{acl}

\usepackage{times}
\usepackage{latexsym}

\usepackage[T1]{fontenc}

\usepackage[utf8]{inputenc}

\usepackage{microtype}
\usepackage{graphicx}
\usepackage{multirow}
\usepackage{amsmath,amsfonts,amssymb}
\usepackage{booktabs}

\usepackage[frozencache=true,cachedir=minted-cache]{minted} 

\usepackage{xcolor}
\usepackage{array}
\usepackage{makecell}
\usepackage[export]{adjustbox}
\usepackage{algorithm}

\title{Fine-grained Image Captioning with CLIP Reward}

\author{
  Jaemin Cho$^1$ \quad
  Seunghyun Yoon$^2$ \quad
  Ajinkya Kale$^3$ \quad
  Franck Dernoncourt$^2$ \quad \\
  \textbf{Trung Bui}$^2$ \quad
  \textbf{Mohit Bansal}$^1$ \\
  $^1$UNC Chapel Hill \quad $^2$Adobe Research \quad $^3$Adobe Inc. \\
  {\tt\small \{jmincho, mbansal\}@cs.unc.edu} \quad
  {\tt\small \{syoon, akale, franck.dernoncourt, bui\}@adobe.com} \\
}

\newcommand{\fgeval}[1]{FineCapEval}

\begin{document}
\maketitle

\input{00_abstact}

\input{01_intro}
\input{02_related_works}
\input{03_methods}

\input{04_dataset}
\input{05_experiments}
\input{06_conclusion}

\bibliography{references}
\bibliographystyle{acl_natbib}

\input{00_appendix}

\end{document}

%% file: 00_abstact.tex
\begin{abstract}

Modern image captioning models are usually trained with text similarity objectives.
However, since reference captions in public datasets often describe the most salient common objects,
models trained with text similarity objectives tend to ignore specific and detailed aspects of an image that distinguish it from others.
Toward more descriptive and distinctive caption generation,
we propose using CLIP, a multimodal encoder trained on huge image-text pairs from web, to calculate multimodal similarity and use it as a reward function.
We also propose a simple finetuning strategy of the CLIP text encoder to improve grammar that does not require extra text annotation.
This completely eliminates the need for reference captions during the reward computation.
To comprehensively evaluate descriptive captions, we introduce \fgeval{}, a new dataset for caption evaluation with fine-grained criteria: overall, background, object, relations.
In our experiments on
text-to-image retrieval and \fgeval{},
the proposed CLIP-guided model generates more distinctive captions than the CIDEr-optimized model.
We also show that our unsupervised grammar finetuning of the CLIP text encoder alleviates the degeneration problem of the naive CLIP reward.
Lastly, we show human analysis where the annotators strongly prefer the CLIP reward
to the CIDEr and MLE objectives according to various criteria.\footnote{
Code and Data: \url{https://github.com/j-min/CLIP-Caption-Reward}}
\end{abstract}

%% file: 01_intro.tex
\section{Introduction}

Describing an image with its detailed and distinguishing aspects is crucial for many applications, such as 
creating text keys for the image search engine and accessibility for the visually impaired.
Standard deep learning approaches train an image-conditioned language model by
maximizing the textual similarity between generated and reference captions~\cite{Vinyals2015b,Xu2015,Rennie2017,Anderson2018}.
However, the reference captions of public datasets often describe only the most prominent objects in the images.
This makes models trained to maximize textual similarity with reference captions tend to generate less distinctive captions that ignore the fine detailed aspects of an image that distinguishes it from others.

\input{fig_method}

To alleviate the problem, we propose to use CLIP~\cite{Radford2021CLIP}, a multimodal encoder model trained on large image-text data (mostly English) collected from the web, by using its similarity scores as rewards (Sec.~\ref{subsec:clip-reward}). 
In addition,
we propose a CLIP text encoder finetuning strategy with synthetic negative caption augmentation
to improve the grammar of captioning model, without any extra text annotations (Sec.~\ref{subsec:clip-finetune}).
Note that our approach completely eliminates the need for reference captions during reward computation.
We illustrate our approach at Fig.~\ref{fig:teaser_singlecolumn}.
To comprehensively evaluate descriptive captions,
we also introduce \fgeval{}, a new dataset that measures captioning in
diverse aspects: overall, background, object, and relation between objects (Sec.~\ref{sec:dataset}).

In our experiments on the MS COCO~\cite{Lin2014COCO} dataset,
we show that the captions of models trained with CLIP reward are more distinctive and contain more detailed information compared to the captions from CIDEr~\cite{vedantam2015cider}-optimized models.  CLIP-guided captions even achieve higher text-to-image retrieval performance than reference captions that are originally paired with images.
We also show that our text encoder finetuning significantly improves caption grammars by removing degeneration artifacts such as word repetition.
In fine-grained caption evaluation with \fgeval{} and human analysis,
we show that our CLIP-based rewards outperform text similarity objectives by a large margin in all categories.

%% file: fig_method.tex
\begin{figure}[t]
    \centering
    \includegraphics[
                    width=\columnwidth,
                    ]{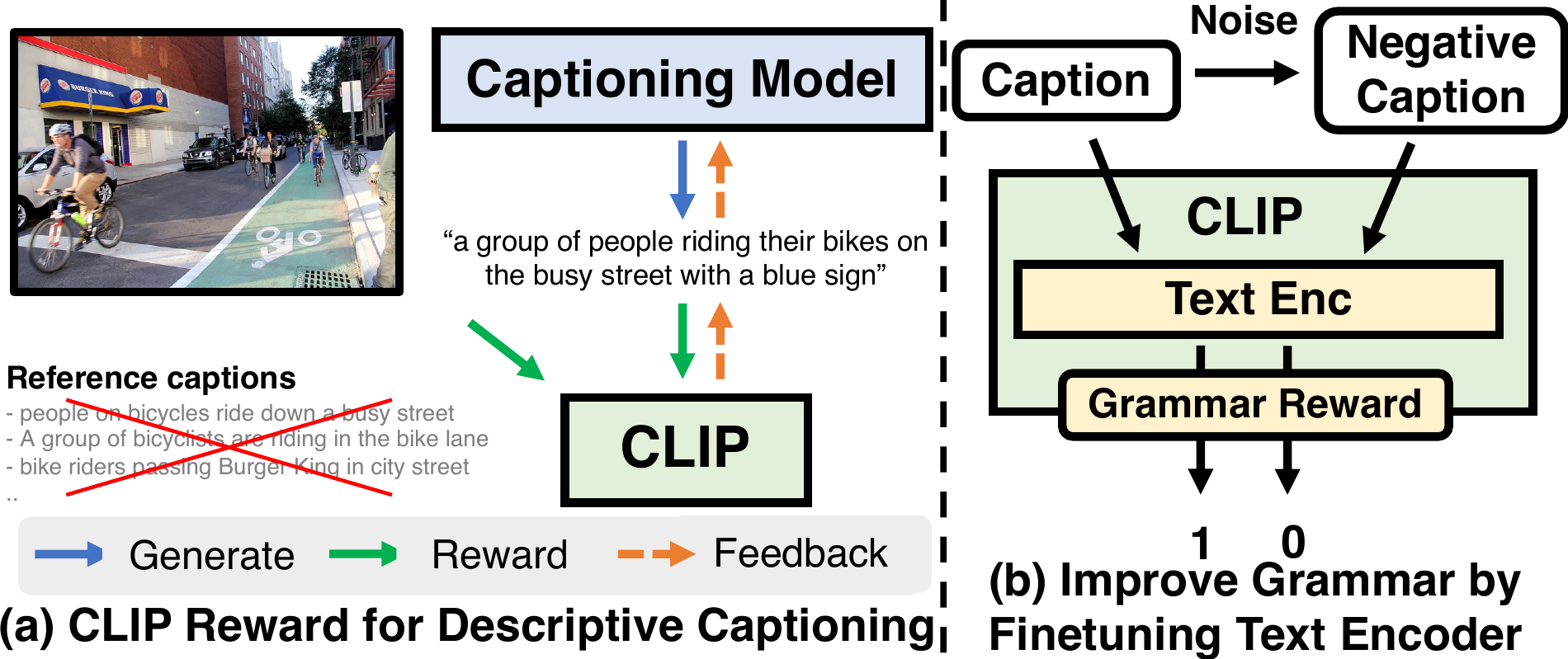}
    \caption{
    Overview of our proposed image captioning method.
    (\textbf{a}) illustrates model training with CLIP based reward that guides descriptive captioning beyond reference captions
    (Sec.~\ref{subsec:clip-reward}).
    (\textbf{b}) shows finetuning CLIP text encoder for improving grammar (Sec.~\ref{subsec:clip-finetune}).
    }
    \label{fig:teaser_singlecolumn}
    \vspace{-10pt}
\end{figure}

%% file: 02_related_works.tex
\section{Related Works}

\paragraph{Image Captioning Metrics.}
Traditionally, captions have been evaluated with similarity metrics based on n-grams or scene graphs, such as
BLEU \cite{Papineni2002},
ROUGE \cite{Lin2004},
METEOR \cite{Banerjee2005},
CIDEr \cite{vedantam2015cider},
and
SPICE \cite{Anderson2016}.
However, such metrics often fail to capture paraphrased expressions due to the limited number of reference captions or scene-graphs.
To address the problem, recent works including
BERTScore \cite{Zhang2019j},
ViLBERTScore \cite{Lee2020vilbertscore},
UMIC \cite{Lee2021},
and
CLIPScore \cite{Hessel2021},
propose using relevance scores computed by language or multimodal models pretrained on large data.

\paragraph{Objectives for Image Captioning.}

Standard deep learning-based image captioning approaches train models with a maximum likelihood estimation (MLE) objective.
\citet{Ranzato2016} point that MLE suffers from an exposure bias problem.\footnote{While language models are trained with ground-truth previous context, they generate words based on the context words previously generated by themselves during inference.}
To address exposure bias,
\citet{Bengio2015} propose a curriculum learning strategy called scheduled sampling.
\citet{Ranzato2016} propose to train models by directly maximizing the text similarity between the generated and reference captions with REINFORCE \cite{Williams1992}.
\citet{Rennie2017,Luo2020} propose self-critical sequence training (SCST) approach by normalizing rewards to stabilize the high variance of rewards.

As illustrated in Fig.~\ref{fig:comparison}, de facto standard reward function for captioning is text similarity between generated and reference captions.
Recent studies have found that reference-trained captioning models often neglect important information from images \cite{Dai2017,Wang2017}.
\citet{Lee2020} use accuracy of an visual question answering model as a reward, encouraging models to generate captions that include information sufficient to answer a visual question.
\citet{Dai2017b,Luo2018,Liu2018} use image-text retrieval model's self-retrieval score as a reward and combine them with metrics based on n-grams, encouraging captioning models to generate captions that are distinctive to each input image.

Note that these works require a careful balance between self-retrieval and text similarity objectives for stable training.
In contrast, with the CLIP text encoder finetuning
(Sec.~\ref{subsec:clip-finetune}), our approach eliminates the need for reference caption and text similarity metrics for the reward computation.

\input{fig_comparison}

%% file: fig_comparison.tex
\begin{figure}[t]
    \centering
    \includegraphics[
                    width=\columnwidth,
                    ]{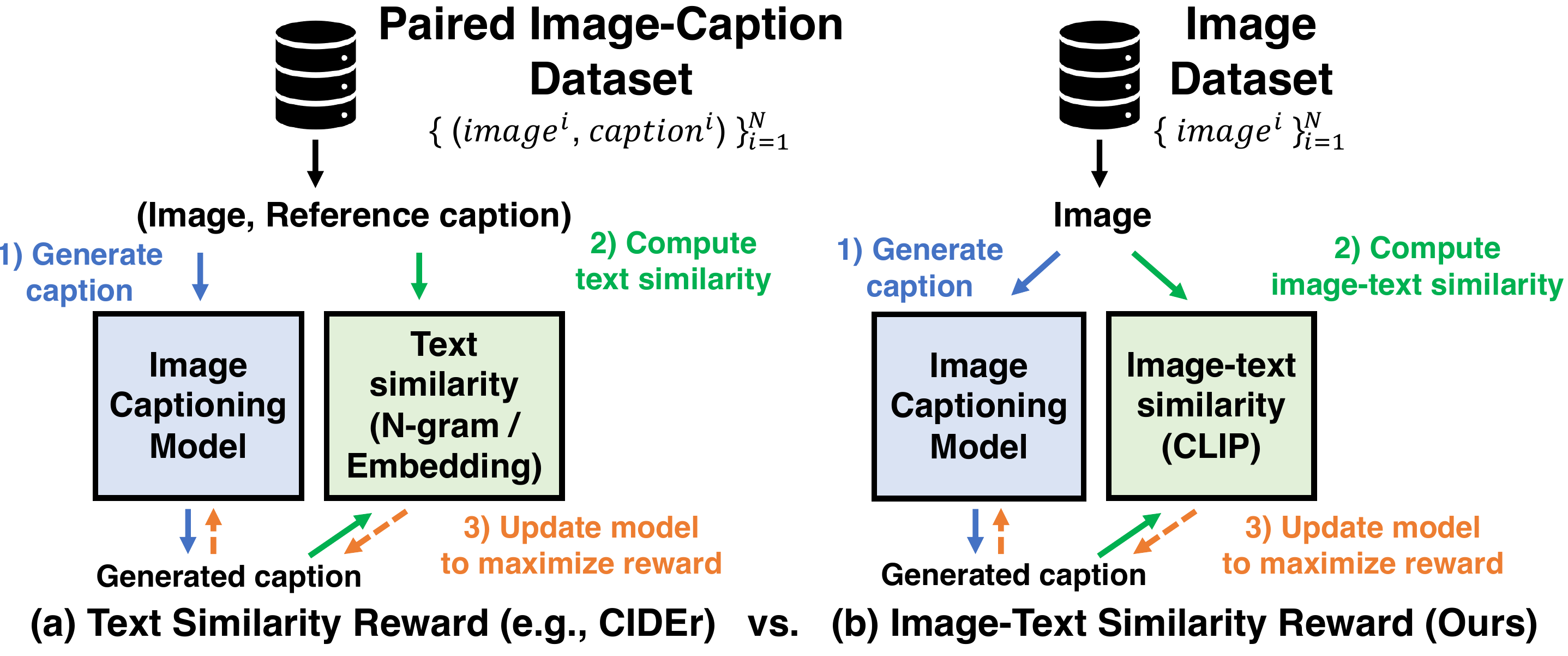}
    \caption{
    Comparison of different reward types for image captioning:
    \textbf{(a)} previous approaches with text similarity reward,
    such as CIDEr~\cite{vedantam2015cider};
    \textbf{(b)} our image-text similarity reward based on CLIP.
    }
    \label{fig:comparison}
    \vspace{-10pt}
\end{figure}

%% file: 03_methods.tex
\input{table_dataset_examples}

\section{Methods}

\subsection{CLIP-guided Image Captioning}
\label{subsec:clip-reward}
We propose using the CLIP \cite{Radford2021CLIP} image-text similarity score to guide a image captioning model. Following \citet{Hessel2021}, we use CLIP-S as our reward:
$\texttt{CLIP-S}(I, c) = w * \max(\frac{f^{I}(I) ^{\intercal} f^{T}(c)}{|f^{I}(I)|\cdot|f^{T}(c)|}, 0)$,
where $I$, $c$ are the image and caption, $f^{I}$, $f^{T}$ are the CLIP image and text encoders, and $w=2.5$.
By learning to maximize the image-text similarity of the contrastive model, image captioning models are encouraged to generate captions that contain more distinctive information about the input image.
Fig.~\ref{fig:teaser_singlecolumn} (a) illustrates this training strategy.

Following \citet{Rennie2017}, we optimize our captioning model $P_\theta(c|I)$ with REINFORCE \cite{Williams1992} with a self-critical baseline.
We approximate the gradient of expected reward for the generated caption $\hat{c}$, where the reward of the beam search is normalized with the baseline reward $b$ from greedy decoding $\hat{c}_{greedy}$: $\nabla_{\theta} \mathop{\mathbb{E}}_{\hat{c} \sim P_\theta(c|I)} [R(I, \hat{c})] \approx (R(I, \hat{c}_{beam}) - R(I, \hat{c}_{greedy})) \nabla_{\theta} \log P_\theta(\hat{c}_{beam}|I)$,
where $R(I, c) = \texttt{CLIP-S}(I, c)$.

\subsection{Improving Grammar with CLIP Text Encoder Finetuning}
\label{subsec:clip-finetune}

Since CLIP is not trained with a language modeling objective, the captioning model trained with the CLIP-S reward often generates grammatically incorrect captions (e.g., repeated words; see Table~\ref{tab:samples}).
We inject grammatical knowledge into the CLIP text encoder with negative captions, generated by randomly repeating/removing/inserting/swapping/shuffling tokens of the reference captions.
We provide the implementation details of such operations in the appendix.
We introduce a grammar head, a two-layer perceptron that takes the CLIP text feature $f^{T}(c)$ as input and produces the probability that $c$ is grammatically correct: $g(c)\in[0,1]$. 
We use binary cross-entropy for the grammar objective, whose label $y$ is 1 for reference captions and 0 for negative captions: $-y \log g(c)$.
We jointly finetune the text encoder and grammar head with the summation of the original CLIP objective and the grammar objective.
Note that we fix the CLIP image encoder parameters during finetuning.
We illustrate the finetuning process in Fig.~\ref{fig:teaser_singlecolumn} (b).
After finetuning CLIP, we train captioning models with rewards augmented with grammar score:
$R(I, c) = \texttt{CLIP-S}(I, c) * \lambda + g(c)$, where $\lambda=2$.

%% file: table_dataset_examples.tex
\begin{table*}[!h]
\centering
\resizebox{\textwidth}{!}{
\begin{tabular}{c l l l}
\toprule
 & Image & Criteria & Annotations \\
\midrule
\multirow{7}[7]{*}{(a)} & \multirow{7}[9]{*}{\includegraphics[width=0.2\textwidth,height=0.2\textwidth,valign=c]{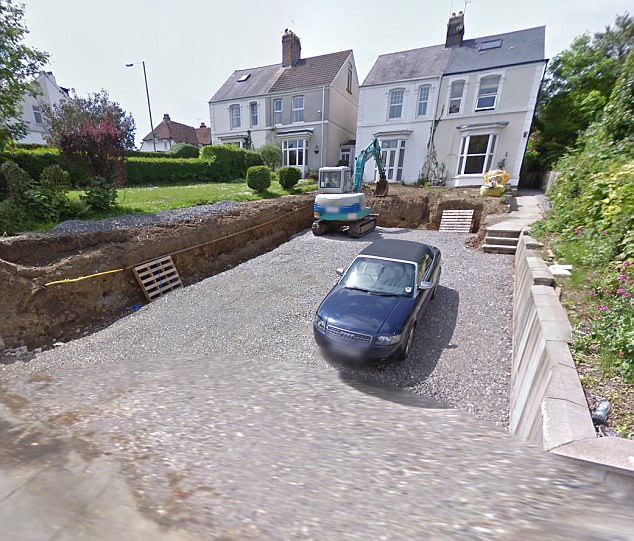}} &
Background & \makecell[l]{white house, truck digging soil in front of the house,
trees and bushes, house surrounded by a small garden,
Mini excavator, houses,\\
white and grey building, greenery,
two houses, blue and white colored machine} \\
\cmidrule{3-4}
& & Object & a blue car,
a blue car,
black car,
car, dozer, white and grey building, greenery,
black car, green bushes \\
\cmidrule{3-4}
& & Relation &  parked in the front yard,
in front, 
parked in front of,
Parked,
car standing on the road \\
\cmidrule{3-4}
& & Overall & \makecell[l]{A blue car parked in the front yard of an off white house with a truck digging soil in front of the house.
\\
A blue car in front of a house surrounded by a small garden with trees and bushes in the background.
\\
A black car parked in front of a house with a mini excavator behind it with other houses in the background.
\\
A car and a dozer parked in front of two white and grey buildings and greenery on both sides.
\\
A black car standing on the road surrounded by green bushes on both sides and two houses and a blue and white colored machine in the background.
} \\
\midrule

\multirow{7}[7]{*}{(b)} &
\multirow{7}[9]{*}{\includegraphics[width=0.2\textwidth,height=0.2\textwidth,valign=c]{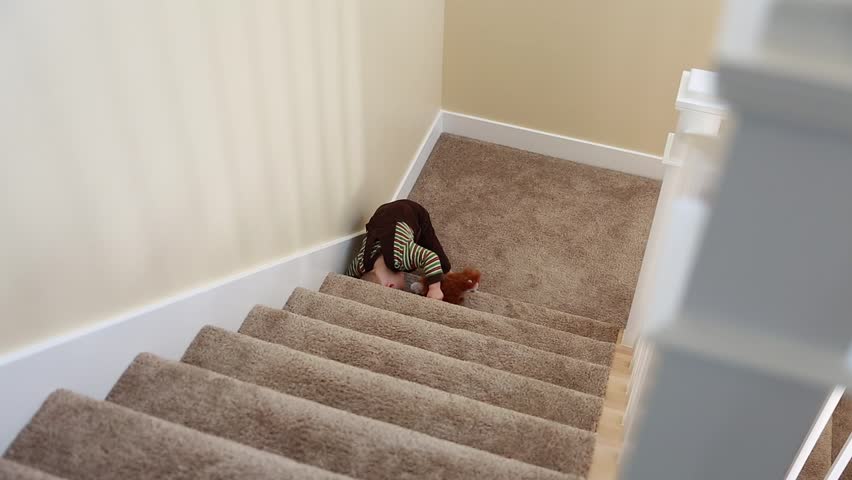}} &
Background & velvet carpet stairs,
light-brown colored stairs, 
Off white wall, 
Cream painted walls,
cream wall with straight line light \\
\cmidrule{3-4}
& & Object & brown jumpsuit, kid,
Toy, black jumpsuit,
boy, brown clothes, toy, brown carpet,
Little young boy, cotton carpeted stair, dark brown jumper dress, cream wall \\
\cmidrule{3-4}
& & Relation & with its head on to,
touching,
Hiding, Holding,
boy holding and playing with the toy,
putting, wearing \\
\cmidrule{3-4}
& & Overall & \makecell[l]{A child wearing a brown jumpsuit with its head on to the velvet carpet stairs. \\
A kid is touching their head on a light brown colored stairs.\\
A Kid wearing a black jumpsuit and holding a toy hiding below the stairs with off white wall in the background. \\
A boy wearing brown clothes holding and playing with his toy and playing on a brown carpet  on stairs with cream painted walls. \\
Little young boy is putting his forehead on the cotton carpeted stair wearing dark brown jumper dress and background of cream wall with straight line light.
} \\

\bottomrule
\end{tabular}
}
\caption{
\fgeval{} examples. For each image, we aggregate the annotations for each criteria from 5 different human annotators. For `overall' criterion, we evaluate captions with CIDEr. For the rest of criteria, we evaluate captions with word-level recall R$_{word}$.
}
\label{tab:fine_grained_dataset_samples}
\end{table*}

%% file: 04_dataset.tex
\section{\fgeval{}: Fine-grained Caption Evaluation Dataset}
\label{sec:dataset}

We introduce \fgeval{}, a new dataset for caption evaluation in four different aspects.
To construct \fgeval{}, we collect 500 images from the MS COCO~\cite{Lin2014COCO} test2015 split and Conceptual Caption~\cite{Sharma2018ConceptualCap} val split, respectively.
Then, for each image, we ask 5 human annotators to write phrases of
1) background,
2) objects (and their attributes; i.e., color, shape, etc.),
3) relation between objects (i.e., spatial relation),
and 
4) a detailed caption that includes all three aspects.
See details of data collection process in appendix.
In total, \fgeval{} consists of 1,000 images with 5,000 annotations for each of the four criteria.
In Table~\ref{tab:fine_grained_dataset_samples}, we show samples from the \fgeval{} dataset.

%% file: 05_experiments.tex
\input{table_karpathy_test}

\section{Experiments}
\label{sec:experiments}

We compare different reward configurations: MLE, CIDEr, CLIP-S, CIDER+CLIP-S, and CLIP-S+Grammar.
Following previous work, we conduct experiments on the MS COCO \cite{Lin2014COCO} English captioning dataset with Karpathy split \cite{Karpathy2015}.
We evaluate the model with n-gram based metrics, embedding based metrics, text-to-image retrieval scores, and \fgeval{}.
We also perform a human evaluation with five criteria to understand the human preference for the generated captions in various aspects.

\paragraph{Model Architecture and Training.}
We use CLIP-Res50$_{\text{Transformer}}$ \cite{Shen2021} as our captioning model architecture.
The model consists of CLIP-Res50 for visual feature extraction and a transformer~\cite{Vaswani2017} encoder-decoder for conditional language model.
We resize images in 224x224 to extract 2048-dimensional visual features.
The transformer consists of 6-layer encoder and 6-layer decoder.
We train our model with MLE objective for 15 epochs and further train with different rewards for 25 epochs (total 40 epochs), which takes within 1 day with 8 V100 GPUs.
We use beam size 5 for beam search decoding. 
We implement a training pipeline with PyTorch~\cite{Paszke2017}, PyTorch Ligthning\footnote{\url{https://github.com/PyTorchLightning/pytorch-lightning}}, and HuggingFace Transformers ~\cite{Wolf2019}.

\paragraph{N-gram based Metrics.}
For N-gram based metrics, we report
BLEU-4 \cite{Papineni2002},
CIDEr \cite{vedantam2015cider}
METEOR \cite{Banerjee2005}, and
ROUGE-L \cite{Lin2004}.

\paragraph{Embedding-based Metrics.}
We report BERT-S \cite{Zhang2019j} and CLIP-S/RefCLIP-S \cite{Hessel2021}.\footnote{Following the default settings of original papers, BERT-S and CLIP-S/RefCLIP-S are based on RoBERTa-Large \cite{Liu2019c} and CLIP ViT-B/32 \cite{Radford2021CLIP} respectively.}
BERT-S measures textual similarity between reference captions and generated captions, CLIP-S measures image-text similarity between input images and generated captions, and RefCLIP-S averages textual similarity (with reference captions) and image-text similarity.

\paragraph{Text-to-Image Retrieval.}
We report the recall of the reference image using a text-to-image retrieval model, to measure the distinctiveness of the generated captions.
For the retrieval model, we use CLIP ViT-B/32 \cite{Radford2021CLIP}.

\paragraph{\fgeval{}.}
For background, object, and relation criteria, we measure caption performance with word-level recall, R$_{word}$ $\in [0,1]$.
See details of R$_{word}$ calculation in appendix.
For overall caption, we measure the performance with CIDEr.

\paragraph{Human Evaluation.}
To evaluate the captions in terms of human preference,
we show a pair of captions from CLIP-S+grammar reward (ours) with CIDEr reward and with MLE baseline to human annotators from Amazon Mechanical Turk\footnote{\url{https://www.mturk.com/}}. Then we ask them to select a better caption on five criteria (overall, background, object, attribute, relation).
For each of the five criteria, we ask 10 annotators with 50 pairwise selection questions.
We use 50 images from \fgeval{} for caption generation.

\input{table_samples}

\section{Results and Discussions}

\subsection{CLIP Guides Distinctive Captions}
In Table~\ref{tab:karpathy_test_metric}, the models with CLIP-S and CLIP-S+Grammar rewards achieve higher image-text metrics (CLIP-S / RefCLIP-S) and text-to-image retrieval scores than baselines.
Interestingly, their retrieval scores are even higher than the retrieval score with reference captions.
This shows the distinctiveness of their generated captions.
For image (a) in Table~\ref{tab:samples}, our model with CLIP-S+Grammar reward describes the rainy weather with `wet', while the model with CIDEr reward does not describe it.

Our models with CLIP-S and CLIP-S+Grammar rewards score lower text similarity metrics (n-gram metrics and BERT-S) than the model with CIDEr reward.
However, the low scores on these reference-based metrics can be addressed by that models with CLIP-S and CLIP-S+Grammar rewards often generate captions that include fine-grained information that is not even present in the reference captions.
For image (b) in Table~\ref{tab:samples}, CLIP-S+Grammar model describes `blue sign' of the restaurant, whereas none of the reference captions mentions them.

\subsection{Finetuning CLIP Text Encoder Improves Grammar}
Table~\ref{tab:samples} shows that the degeneration (e.g. repetition of words) of the CLIP-S reward is successfully mitigated by adding the grammar reward (CLIP-S+Grammar).
Table~\ref{tab:karpathy_test_metric} shows that adding grammar reward significantly increases all text similarity metrics (e.g., +60 for CIDEr).

\input{table_amt_pairwise}

\subsection{Fine-grained Caption Evaluation}

\paragraph{\fgeval{}.}
The four right columns of
Table~\ref{tab:karpathy_test_metric} show that
CLIP-S and CLIP-S+Grammar significantly outperform CIDEr on all four criteria of \fgeval{}: overall, background, object, relation.
The gap is smallest in the object criterion, which implies that MS COCO reference captions describe more object information than background or relation between objects.

\paragraph{Human Evaluation.}
Table~\ref{tab:amt_pairwise} shows human evaluation results
on five criteria: overall, background, object, attribute, relation.
We sample 50 captions from model trained with CLIP-S+grammar reward (ours), CIDEr reward and MLE baseline using 50 images from Conceptual caption~\cite{Sharma2018ConceptualCap} val split.
For each of the five criteria, we ask 10 human annotators to select a better caption between ours and another method.
On all criteria, human annotators strongly prefer captions with CLIP-S+Grammar rewards over CIDEr and MLE baseline.

%% file: table_karpathy_test.tex
\begin{table*}[t]
\centering
\resizebox{\textwidth}{!}{
\begin{tabular}{l  cccc  ccc | ccc | cccc}
\toprule
\multirow{3}[3]{*}{Reward} & \multicolumn{4}{c|}{N-gram based}  & \multicolumn{3}{c|}{Embed based} & \multicolumn{3}{c|}{\multirow{2}[2]{*}{Text-to-Image Retrieval}} & \multicolumn{4}{c}{\fgeval{}} \\
\cmidrule{2-5} \cmidrule{6-8}
& \multicolumn{5}{c|}{Text based} & \multicolumn{2}{c|}{Image-Text based} & & & & Overall & Bg. & Obj. & Rel.  \\
\cmidrule{2-6} \cmidrule{7-8}
& BLEU-4 & CIDEr & METEOR & ROUGE-L & BERT-S & CLIP-S & RefCLIP-S & R@1 & R@5 & R@10 & CIDEr & R$_{word}$ & R$_{word}$ & R$_{word}$\\
\midrule

Reference captions & & & & & & & & 29.5$^{\star}$ & 54.2$^{\star}$ & 65.0$^{\star}$ \\
\midrule
MLE & 32.5 & 110.3 & 27.2 & 55.2 & 0.937 & 0.758 & 1.12 & 21.8 & 45.6 & 58.0 & 13.5 & 11.6 & 13.0 & 19.8 \\
CIDEr & \textbf{38.2} & \textbf{124.9} & 28.7 & \textbf{58.5} & \textbf{0.942} & 0.759 & 1.13 & 20.9 & 45.6 & 58.2 & 12.8 & 13.1 & 23.1 & 22.4\\
CLIP-S & 6.2 & 11.2 & 18.7 & 31.6 & 0.882 & \textbf{0.860} & \textbf{1.17} & \textbf{42.5} & \textbf{71.6} & \textbf{82.2} & 13.9 & 20.8 & \textbf{26.4} & 24.9 \\
CIDEr+CLIP-S & 37.7 & 124.6 & \textbf{28.8} & 58.3 & 0.941 & 0.772 & 1.14 & 24.4 & 50.2 & 63.1 & 13.0 & 13.0 & 23.4 & 21.7 \\
CLIP-S+Grammar & 16.9 & 71.0 & 24.9 & 47.3 & 0.924 & 0.793 & 1.15 & 35.8 & 64.0 & 75.8 & \textbf{19.3} & \textbf{21.8} & 25.5 & \textbf{27.5} \\
\bottomrule
\end{tabular}
}
\caption{
Captioning performance of different rewards on MS COCO Karpathy test split.
$^{\star}$The first caption out of 5 reference captions is used to calculate retrieval scores.
R@K refers to the recall-K of the reference image.
R$_{word}$ refers to the word-level recall for background (Bg.), object (Obj.) and relation (Rel.) criteria (see Sec.~\ref{sec:dataset} for details).
}
\label{tab:karpathy_test_metric}
\end{table*}

%% file: table_samples.tex
\begin{table*}[!h]
\centering
\resizebox{\textwidth}{!}{
\begin{tabular}{c l l l}
\toprule
 & Image & Reward & Captions \\
\midrule

\multirow{7}[7]{*}{(a)} &
\multirow{7}[9]{*}{\includegraphics[width=0.2\textwidth,height=0.2\textwidth,valign=c]{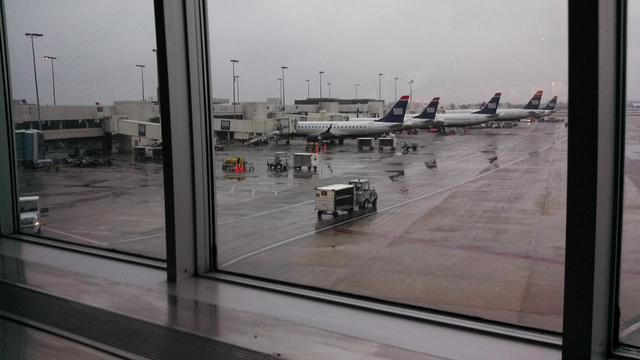}} &
CIDEr & a window of an airport with planes on the runway \\
\cmidrule{3-4}
& & CLIP-S & several rows of planes parked outside a terminal window area with fog outside a terminal window motion position area motionn \\
\cmidrule{3-4}
& & CLIP-S + Grammar & a lot of airplanes parked on a wet airport terminal \\
\cmidrule{3-4}
& & Reference Captions & \makecell[l]{An airport filled with planes sitting on tarmacs.
\\
The view of runway from behind the windows of airport.
\\
a truck driving towards some planes parked on the runway
\\
Planes on a wet tarmac unloading at arrival gates.
\\
Window view from the inside of airplanes, baggage carrier and tarmac.
} \\
\midrule
\multirow{7}[7]{*}{(b)} &
\multirow{7}[9]{*}{\includegraphics[width=0.2\textwidth,height=0.2\textwidth,valign=c]{images/COCO_val2014_000000462565.jpeg}} &
CIDEr & a group of people riding bikes down a city street \\
\cmidrule{3-4}
& & CLIP-S & several cyclists moving and bicycles near a restaurant and a blue advertisement outside a red brick building motion stance p \\
\cmidrule{3-4}
& & CLIP-S + Grammar & a group of people riding their bikes on the busy street with a blue sign \\
\cmidrule{3-4}
& & Reference Captions & \makecell[l]{people on bicycles ride down a busy street
\\
A group of people are riding bikes down the street in a bike lane
\\
bike riders passing Burger King in city street
\\
A group of bicyclists are riding in the bike lane.
\\
Bicyclists on a city street, most not using the bike lane
} \\

\bottomrule
\end{tabular}
}
\caption{
Captions generated by models with different rewards on MS COCO Karpathy test split images.
}
\label{tab:samples}
\end{table*}

%% file: table_amt_pairwise.tex
\begin{table}[]
\centering
\resizebox{\columnwidth}{!}{
\begin{tabular}{l | l | ccc}
\toprule
Criteria &  CLIP-S + Grammar & Win & Lose & Tie  \\
\midrule
\multirow{2}{*}{Overall} & v.s. MLE & \textbf{49.0} & 41.8 & 9.2 \\
& v.s. CIDEr & \textbf{51.0} & 30.8 & 18.2 \\
\midrule
\multirow{2}{*}{Background} & v.s. MLE  & \textbf{52.8} & 35.0 & 12.2 \\
& v.s. CIDEr & \textbf{53.9} & 25.4 & 20.6 \\
\midrule
\multirow{2}{*}{Object} & v.s. MLE  & \textbf{52.0} & 36.6 & 11.4 \\
& v.s. CIDEr & \textbf{55.2} & 32.8 & 12.0 \\
\midrule
\multirow{2}{*}{Attribute} & v.s. MLE  & \textbf{57.2} & 36.8 & 6.0 \\
& v.s. CIDEr & \textbf{55.8} & 37.2 & 7.0 \\
\midrule
\multirow{2}{*}{Relation} & v.s. MLE  & \textbf{44.6} & 44.2 & 11.2 \\
& v.s. CIDEr & \textbf{49.2} & 39.6 & 11.2  \\
\bottomrule
\end{tabular}
}
\caption{
Human pairwise preference evaluation results.
}
\label{tab:amt_pairwise}
\vspace{-10pt}
\end{table}

%% file: 06_conclusion.tex
\section{Conclusion and Future Directions}
We introduce a novel training strategy for image captioning models by maximizing multimodal similarity score of CLIP and finetuning its text encoder to improve grammar.
The use of CLIP reward eliminates the need for reference captions and their bias for the reward computation.
We also introduce \fgeval{}, a dataset for fine-grained caption evaluation.
We demonstrate the effectiveness of our proposed method based on
improvements in text-to-image retrieval, \fgeval{}, and human evaluation on fine-grained criteria along with qualitative examples.
Future works involve
finetuning CLIP reward models with desired writing styles for different applications
and improving the synthetic augmentation process by using external data suitable for grammars with advanced linguistics expertise.

\section{Ethical Considerations}
The CLIP models that we used are trained on millions of web  image-text pairs. \citet{Birhane2021} shows that such large-scale datasets often contain explicit and problematic image-text pairs.
As the CLIP model card\footnote{\url{https://github.com/openai/CLIP/blob/main/model-card.md}} suggests, the use of CLIP reward to train image captioning models is intended as a research output, and any deployed use case of the models is out of scope.

Our captioning models and CLIP models are trained on English datasets; its use should be limited to English language use cases.
As our proposed method is not limited to English and is easily extended to other languages, future work will explore the extensions in various languages.

\section*{Acknowledgements}
We thank the reviewers for their valuable comments. This work was partially done while JC was interning at Adobe Research and later extended at UNC, where it was supported by ARO Award W911NF2110220, DARPA MCS Grant N66001-19-2-4031, and NSF-CAREER Award 1846185. The views contained in this article are those of the authors and not of the funding agency.

%% file: 00_appendix.tex
\appendix

\input{table_samples_full}

In this appendix,
we include more example image captioning with different rewards (Sec.~\ref{sec:caption_examples}),
implementation details (Sec.~\ref{sec:exp_details}),
\fgeval{} details (Sec.~\ref{sec:dataset_details}),
human evaluation details (Sec.~\ref{sec:human_eval_details}),
and the license for the datasets and models used in this project (Sec.~\ref{sec:license}).

\section{More Image Captioning Examples}
\label{sec:caption_examples}
We provide more image captioning examples using different reward functions in Table~\ref{tab:samples_full}. Overall, the captions from the model with CLIP-S+Grammar reward provide 1) more descriptive than the captions from the CIDEr model and reference captions, and 2) more grammatically correct than the captions from the model with CLIP-S reward.

\section{Implementation Details}
\label{sec:exp_details}

\paragraph{Negative Caption Generation.}
In Alg.~\ref{alg:neg_generation}, we show Python implementation of the negative text generation (Sec.~\ref{subsec:clip-finetune}) for grammar finetuning.
In summary, we generate negative captions using one of the operations: \texttt{repeat}, \texttt{remove}, \texttt{insert}, \texttt{swap}, \texttt{shuffle} on the original captions.

\paragraph{Evaluation Scripts.}
We use pycocoevalcap\footnote{\url{https://github.com/tylin/coco-caption/tree/master/pycocoevalcap}} for MS COCO caption evaluation metrics such as CIDEr.
We use BERTScore official repo\footnote{\url{https://github.com/Tiiiger/bert_score}} with \texttt{roberta-large} model to calculate BERT-S.
We report the evaluation script number from single run (single weight initialization), as we did not observe meaningful score fluctuation across multiple runs in our initial experiments.

\section{\fgeval{} Details}
\label{sec:dataset_details}

\paragraph{Data Collection.}
\label{sec:details_data_colleciton_process}
To create a fine-grained description of the image, we ask annotators to write a caption that should describe target images' 1) background, 2) objects and their attributes (i.e., color, shape, etc.), and 3) the relationship between the objects if any (i.e., spatial relation). Furthermore, we ask the annotators to write metadata containing which words/phrases in their writing belong to the three criteria. We also provide annotators with guidelines in writing a caption as follows:
1) There should be a single sentence describing the image.
2) The image may be a photo, an illustration or a pure background. 
3) Pay close attention to local and global events in the image.
4) Descriptions should be at least ten words for each image.
5) Avoid the subject description of the image (i.e., a dog runs ``very fast'', a man feels ``successful'').
6) Avoid known entities such as specific locations (i.e. Eifel Tower), time (i.e., 4 pm), event (i.e., Halloween), proper name.
7) In describing people, use only man/woman/boy/girl if clear; otherwise, use person/child.
All annotators are hired by a professional crowdsourcing platform TELUS\footnote{\url{http://www.telusinternational.com}}.
The crowdsourcing company obtained consents from the crowdworkers before the annotation process and conducted the ethical reviews.
We collect English captions and all the annotators are native English speakers living in the US.
We pay 5,400 USD, including 1) caption creation (5k samples) and 2) quality assurance process that manually examines 50\% of the created caption by different workers.

\paragraph{Word-level Recall R$_{word}$.}
In Alg.~\ref{alg:word_recall}, we show Python implementation of word-level recall R$_{word}$.
In summary, R$_{word}$ measures how many words from each of the reference phrases are included in a generated caption on average.

\input{alg_text_augmentation}
\input{alg_word_recall}

\input{fig_human_eval_screenshot}
\section{Human Evaluation Details}
\label{sec:human_eval_details}
We conduct pairwise evaluation of human preference, as shown in the Sec.~\ref{sec:experiments}.
For each image,
we show two captions generated from two models: ours (CLIP-S + Grammar) and the baseline (MLE/CIDEr).
A human worker selects a caption that better describes the image in terms of five criteria: overall, background, object, attribute, and relation.
For each criterion, we use 50 images from \fgeval{}, and the two options are randomly and evenly shuffled. We also provide `Tie' option to choose when the two captions are equally good or bad.
For each criterion, we recruit 10 annotators 1) who are located in the  Great Britain or the United States 2) HIT approval rate above 80\% and 3) Number of HITs approved greater than 1000, from Amazon Mechanical Turk.
We pay the annotators 0.03 USD per selection, which roughly corresponds to 11 USD/hour.
In Fig.~\ref{fig:human_eval_screenshot}, we provide the screenshot for `object' criterion for example.

\section{Licenses}
\label{sec:license}

For all artifacts, we remain within their respective license agreements. Here, we list the licenses:
\begin{itemize}
    \item MS COCO - CC 4.0 - \url{https://cocodataset.org/#termsofuse}
    \item Conceptual Captions - \url{https://github.com/google-research-datasets/conceptual-captions/blob/master/LICENSE}
    \item CLIP - MIT - \url{https://github.com/openai/CLIP/blob/main/LICENSE}
    \item CLIP-ViL - MIT -  \url{https://github.com/clip-vil/CLIP-ViL/blob/master/LICENSE}
\end{itemize}

%% file: table_samples_full.tex
\begin{table*}[!h]
\centering
\resizebox{\textwidth}{!}{
\begin{tabular}{c l l l}
\toprule
 & Image & Reward & Captions \\
\midrule
\multirow{7}[7]{*}{(a)} & \multirow{7}[9]{*}{\includegraphics[width=0.2\textwidth,height=0.2\textwidth]{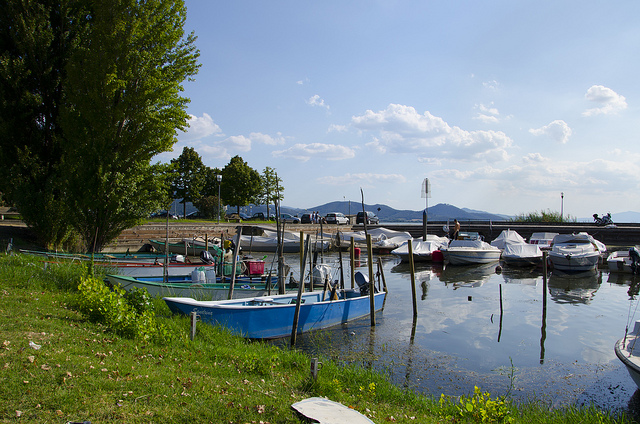}} & CIDEr & a group of boats parked in the water on a lake \\
\cmidrule{3-4}
& & CLIP-S & several rows of boats parked near a canal mountains horizon area and a mountain horizon horizon area horizon ear motion \\
\cmidrule{3-4}
& & CLIP-S+Grammar & a lot of boats parked on the grass next to the lake with the hills behind \\
\cmidrule{3-4}
& & Reference Captions & \makecell[l]{A blue boat docked on a green lush shore. \\
A small marina with boats docked there \\
a group of boats sitting together with no one around \\ 
Some boats parked in the water at a dock \\
boats sitting around the side of a lake by a tree} \\
\midrule

\multirow{7}[7]{*}{(b)} & \multirow{7}[9]{*}{\includegraphics[width=0.2\textwidth,height=0.2\textwidth,valign=c]{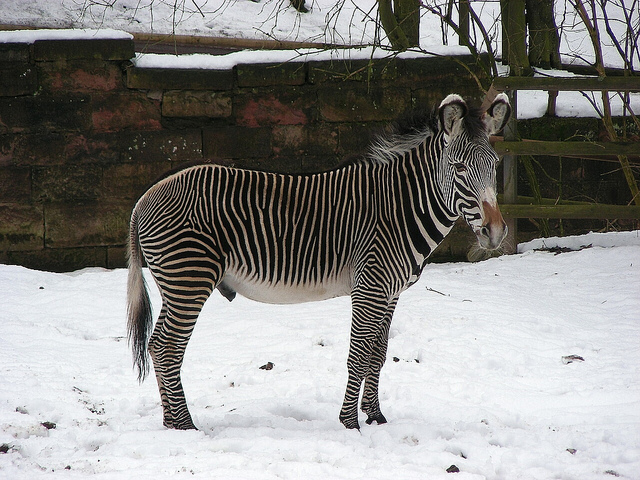}} & CIDEr & a zebra standing in the snow next to a brick wall \\
\cmidrule{3-4}
& & CLIP-S & a adult zebra wearing black and grey stripes standing near a brick wall area area with grey stance position stance \\
\cmidrule{3-4}
& & CLIP-S+Grammar & a large black and grey zebra standing together in the snowy ground next to a stone \\
\cmidrule{3-4}
& & Reference Captions & \makecell[l]{A zebra is standing outside in the snow
\\
One zebra standing in snow near a stone wall.
\\
A zebra is standing in a snowy field.
\\
A zebra stands in snow in front of a wall.
\\
A zebra standing alone in the snow with a stone block wall and wooden fence behind it.
} \\
\midrule

\multirow{7}[7]{*}{(c)} & \multirow{7}[9]{*}{\includegraphics[width=0.2\textwidth,height=0.2\textwidth,valign=c]{images/COCO_val2014_000000288955.jpeg}} & CIDEr & a black dog sitting next to a plate of food \\
\cmidrule{3-4}
& & CLIP-S & black black dog with macaroni and macaroni plate with pasta and pasta on a wooden floor plate position position position \\
\cmidrule{3-4}
& & CLIP-S+Grammar & a black dog sitting next to a plate of food on the wood floor \\
\cmidrule{3-4}
& & Reference Captions & \makecell[l]{Shaggy dog gets dinner served on a plate.
\\
A small black dog standing over a plate of food.
\\
A small dog eating a plate of broccoli.
\\
A black dog being given broccoli to eat.
\\
There is a dog staring at a plate of food} \\
\midrule

\multirow{7}[7]{*}{(d)} & \multirow{7}[9]{*}{\includegraphics[width=0.2\textwidth,height=0.2\textwidth,valign=c]{images/COCO_val2014_000000314294.jpeg}} &
CIDEr & two elephants standing next to a tree in a zoo \\
\cmidrule{3-4}
& & CLIP-S & two adult adult and baby elephant near a tree enclosure area with a tree area enclosure motion stance ear stance \\
\cmidrule{3-4}
& & CLIP-S+Grammar & a large elephant playing with a tree in the dirt field with rocks behind it \\
\cmidrule{3-4}
& & Reference Captions & \makecell[l]{An elephant standing under the shade of a tree.
\\
An elephant standing in the middle of a rocky environment.
\\
An elephant is alone in a wooded enclosure.
\\
An elephant standing in a shaded cleaning in a wooded area.
\\
An elephant walks alone past some big rocks boulders in an open field} \\
\midrule

\multirow{7}[7]{*}{(e)} & \multirow{7}[9]{*}{\includegraphics[width=0.2\textwidth,height=0.2\textwidth,valign=c]{images/COCO_val2014_000000462565.jpeg}} &
CIDEr & a group of people riding bikes down a city street \\
\cmidrule{3-4}
& & CLIP-S & several cyclists moving and bicycles near a restaurant and a blue advertisement outside a red brick building motion stance p \\
\cmidrule{3-4}
& & CLIP-S+Grammar & a group of people riding their bikes on the busy street with a blue sign \\
\cmidrule{3-4}
& & Reference Captions & \makecell[l]{people on bicycles ride down a busy street
\\
A group of people are riding bikes down the street in a bike lane
\\
bike riders passing Burger King in city street
\\
A group of bicyclists are riding in the bike lane.
\\
Bicyclists on a city street, most not using the bike lane
} \\
\midrule

\multirow{7}[7]{*}{(f)} & \multirow{7}[9]{*}{\includegraphics[width=0.2\textwidth,height=0.2\textwidth,valign=c]{images/COCO_val2014_000000483108.jpeg}} &
CIDEr & a man riding a bike next to a train \\
\cmidrule{3-4}
& & CLIP-S & older adult male riding a bicycle near a red and commuter train passing a train station motion stance ear stance \\
\cmidrule{3-4}
& & CLIP-S+Grammar & a person walking on a bike next to a red passenger train on the road \\
\cmidrule{3-4}
& & Reference Captions & \makecell[l]{A man on a bicycle riding next to a train
\\
A person is riding a bicycle but there is a train in the background.
\\
a red and white train and a man riding a bicycle
\\
a guy that is riding his bike next to a train
\\
A man riding a bike past a train traveling along tracks.} \\
\midrule

\multirow{7}[7]{*}{(g)} & \multirow{7}[9]{*}{\includegraphics[width=0.2\textwidth,height=0.2\textwidth,valign=c]{images/COCO_val2014_000000560623.jpeg}} &
CIDEr & a window of an airport with planes on the runway \\
\cmidrule{3-4}
& & CLIP-S & several rows of planes parked outside a terminal window area with fog outside a terminal window motion position area motionn \\
\cmidrule{3-4}
& & CLIP-S+Grammar & a lot of airplanes parked on a wet airport terminal \\
\cmidrule{3-4}
& & Reference Captions & \makecell[l]{An airport filled with planes sitting on tarmacs.
\\
The view of runway from behind the windows of airport.
\\
a truck driving towards some planes parked on the runway
\\
Planes on a wet tarmac unloading at arrival gates.
\\
Window view from the inside of airplanes, baggage carrier and tarmac.
} \\

\bottomrule
\end{tabular}
}
\caption{
More captions generated by models with different rewards on MS COCO Karpathy test split images.
}
\label{tab:samples_full}
\end{table*}

%% file: alg_text_augmentation.tex
\definecolor{bg}{rgb}{0.95,0.95,0.95}

\begin{algorithm*}
\begin{minted}{python}
from random import randint, choice, shuffle
def repeat(tokens, n_max_gram=3, n_max_repeat=3): # repeat n-grams
    n_gram = randint(1, n_max_gram)
    repeat_idx = randint(0, len(tokens) - n_gram)
    repeated = tokens[repeat_idx:repeat_idx+n_gram]
    n_repeat = randint(1, n_max_repeat)
    for _ in range(n_repeat):
        insert_idx = randint(0, len(tokens))
        tokens = tokens[:insert_idx]+repeated+tokens[insert_idx:]
    return tokens
def remove(tokens, n_max_gram=3): # remove n-grams
    n_gram = randint(1, n_max_gram)
    remove_idx = randint(0, len(tokens) - n_gram)
    tokens = tokens[:remove_idx] + tokens[remove_idx + n_gram:]
    return tokens
def insert(tokens, vocab, n_max_tokens=3): # insert tokens
    n_insert_token = randint(1, n_max_tokens)
    for _ in range(n_insert_token):
        insert_idx = randint(0, len(tokens) - 1)
        insert_tok = choice(vocab)
        tokens = tokens[:insert_idx]+[insert_tok]+tokens[insert_idx:]
    return tokens
def swap(tokens, vocab, n_max_tokens=3): # swap tokens
    n_swap_tokens = randint(1, n_max_tokens)
    for _ in range(n_swap_tokens):
        swap_token_idx = randint(0, len(tokens) - 1)
        swap_token = choice(vocab)
        while swap_token == tokens[swap_token_idx]:
            swap_token = choice(vocab)
        tokens[swap_token_idx] = swap_token
    return tokens
def _shuffle(tokens): # shuffle tokens
    shuffle(tokens)
    return tokens
def generate_negative_text(text, vocab): # main function
    tokens = text.split()
    neg_type = choice(['repeat','remove','insert','swap','shuffle'])
    if neg_type == 'repeat': tokens = repeat(tokens)
    elif neg_type == 'remove': tokens = remove(tokens)
    elif neg_type == 'insert': tokens = insert(tokens, vocab)
    elif neg_type == 'swap': tokens = swap(tokens), vocab)
    elif neg_type == 'shuffle': tokens = _shuffle(tokens)
    return " ".join(tokens)
\end{minted}
\caption{Python implementation of negative text generation (main paper Sec.~\ref{subsec:clip-finetune})}
\label{alg:neg_generation}
\end{algorithm*}

%% file: alg_word_recall.tex
\begin{algorithm*}
\begin{minted}{python}
def calculate_word_recall(pred_id2sent, gt_id2phrases):
    """
    pred_id2sent: dict of generated captions (dict[int, str])
    gt_id2phrases: dict of reference phrases (dict[int, list[str]])
    """
    n_total = 0
    total_score = 0
    for id, gt_phrases in gt_id2phrases.items():
        pred_sent = pred_id2sent[id]
        score = 0
        for gt_phrase in gt_phrases:
            word_score = 0
            for gt_word in gt_phrase.split():
                if gt_word in pred_sent:
                    word_score += 1
            score += word_score / len(gt_phrase.split())
        score /= len(gt_phrases)
        total_score += score
        n_total += 1
    word_recall = total_score / n_total * 100
    return word_recall
\end{minted}
\caption{Python implementation of word-level recall R$_{word}$ computation (main paper Sec.~\ref{sec:experiments})}
\label{alg:word_recall}
\end{algorithm*}

%% file: fig_human_eval_screenshot.tex
\begin{figure*}[]
    \centering
    \includegraphics[
                    width=\textwidth,
                    ]{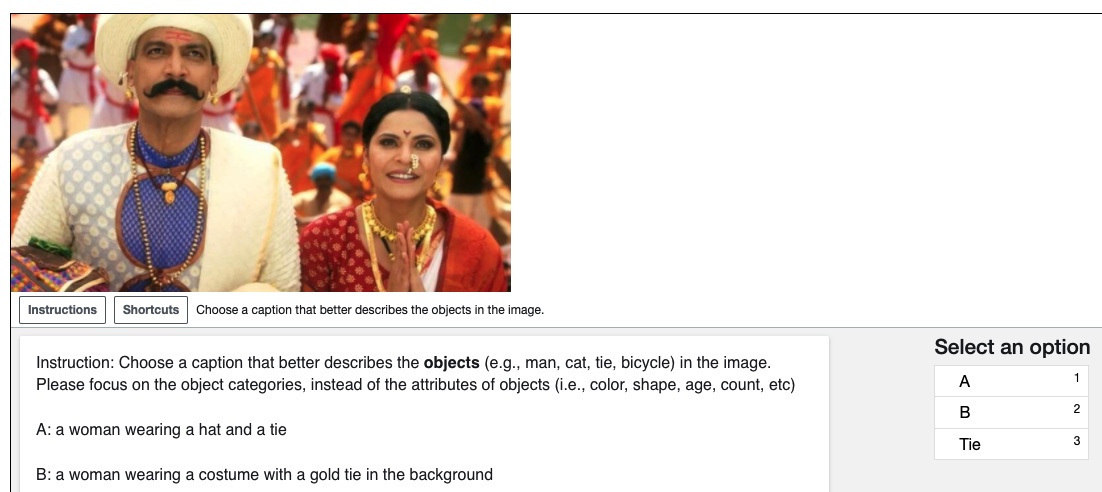}
    \caption{
    The screenshot of human evaluation process for `object' criterion (main paper Sec.~\ref{sec:experiments}).
    }
    \label{fig:human_eval_screenshot}
\end{figure*}